\pdfoutput=1
\documentclass[11pt]{article}

\usepackage[]{emnlp2021}

\usepackage{times}
\usepackage{latexsym}
\usepackage{graphicx}
\usepackage[T1]{fontenc}
\usepackage[utf8]{inputenc}

\usepackage{microtype}

\usepackage{balance}

\newcommand{\nop}[1]{}

\title{Improving Punctuation Restoration for Speech Transcripts via External Data}

\author{Xue-Yong Fu, \ Cheng Chen, \ Md Tahmid Rahman Laskar, \\ {\bf Shashi Bhushan TN, \ Simon Corston-Oliver}\\
        Dialpad Canada Inc.\\ Vancouver, BC, Canada\\
        \texttt{\{xue-yong,cchen,tahmid.rahman\}@dialpad.com}\\\texttt{\{sbhushan,scorston-oliver\}@dialpad.com}}

\begin{document}

\maketitle

\begin{abstract}
Automatic Speech Recognition (ASR) systems generally do not produce punctuated transcripts. To make transcripts more readable and follow the expected input format for downstream language models, it is necessary to add punctuation marks. In this paper, we tackle the punctuation restoration problem specifically for the noisy text (e.g., phone conversation scenarios). To leverage the available written text datasets, we introduce a data sampling technique based on an n-gram language model to sample more training data that are similar to our in-domain data. Moreover, we propose a two-stage fine-tuning approach that utilizes the sampled external data as well as our in-domain dataset for models based on BERT. Extensive experiments show that the proposed approach outperforms the baseline with an improvement of $1.12\%$ F1 score.

% CC's version
\nop{
Transcripts of naturally occurring human-human conversation produced by automatic speech recognition (ASR) systems can be difficult for users to read even if the transcripts have a low error rate. Due to speech disfluency, it becomes more challenging if ASR systems produce transcripts without punctuation. We discuss the problem of editing transcripts to improve readability by adding punctuation in a realtime system. We present a new way of data sampling to obtain external pre-training data that is similar to our in-domain data. We evaluate different training approaches and produce a comprehensive comparison. We show that the model with analogous in-domain pre-training achieve the best results.  
}

\end{abstract}

\section{Introduction}
\label{sec:intro}

%\nop{
%Most ({\color{red} cc: any evidence to support ``most"?}) Automatic Speech Recognition (ASR) systems don’t produce punctuation marks in the output. This significantly reduces readability of transcripts and user experience. It is also very important for downstream natural language processing (NLP) tasks such as sentiment analysis, named entity recognition, aspect mining to consume punctuated texts as punctuation marks provide clearly-defined sentence boundaries.  
%}

%ASR systems have been widely deployed in many products to support a variety of business applications, such as customer support, virtual assistant and conversational AI systems. ASR systems trained on large amounts of data can transcribe human speech with high accuracy~\cite{DBLP:conf/interspeech/KannanDSWRWBCL19, DBLP:conf/nips/BaevskiZMA20}. Recent advancements in ASR made it possible to transcribe conversations for real-time assistance, note-taking or agent training. However, highly accurate transcripts may suffer readability issues along with false starts, repetitions, etc. Since spoken dialogue transcripts produced by ASR systems often do not contain any punctuations, users reading long unpunctuated utterances may find such transcripts unreadable. % distracting. %and unhelpful.

ASR systems are widely deployed in many products to support a variety of business applications, such as customer support, virtual assistants, and conversational systems. The context for this research is to build commercial products that perform realtime transcription of English business telephone calls, with additional natural language processing to enable features such as assistance for customer support agents and note-taking during business meetings. Though ASR systems trained on large amounts of data can transcribe human speech with high accuracy  ~\cite{DBLP:conf/interspeech/KannanDSWRWBCL19, DBLP:conf/nips/BaevskiZMA20}, spoken dialogue transcripts produced by such systems often exclude punctuation marks. Therefore, the user reading a long unpunctuated transcript may find it unreadable \cite{DBLP:conf/interspeech/ZelaskoSMSCD18,DBLP:conf/aclnut/AlamKA20}.

%ASR systems trained on large amounts of data can transcribe human speech with high accuracy  ~\cite{DBLP:conf/interspeech/KannanDSWRWBCL19, DBLP:conf/nips/BaevskiZMA20}. However, spoken dialogue transcripts produced by ASR systems often exclude punctuation marks, and the user reading a long unpunctuated transcript may find it unreadable \cite{DBLP:conf/interspeech/ZelaskoSMSCD18,DBLP:conf/aclnut/AlamKA20}. 

% which also made it possible to transcribe conversations for real-time assistance, note-taking, or agent training

% However, highly accurate transcripts may still suffer readability issues along with false starts, repetitions, and etc. Since spoken dialogue transcripts produced by ASR systems often exclude punctuation marks, the user reading a long unpunctuated transcript may find it unreadable \cite{DBLP:conf/interspeech/ZelaskoSMSCD18,DBLP:conf/aclnut/AlamKA20}.

\nop{Since phone or voice is still the major channel for customers dealing with contact centers, the ability for agents or managers to read transcripts or real-time transcription of calls they make has tremendous business value.}

Beyond the need for readability, downstream natural language understanding (NLU) models use  transcripts to provide more business insights. Many NLU models (e.g., sentiment analysis and named entity recognition (NER)) achieving state-of-the-art performance are based on large-scale language models (e.g. BERT~\cite{DBLP:conf/naacl/DevlinCLT19}), which are often pre-trained on well-punctuated texts such as English Wikipedia and BookCorpus~\cite{DBLP:conf/iccv/ZhuKZSUTF15}. %Punctuation in these datasets can be used to do sentence segmentation %and sentences are common model input units. 
However, NLU models trained on such datasets may not perform well on unpunctuated transcripts. Thus, for the conversational domain, punctuated speech transcripts are also necessary.

\nop{for example, out in-domain utterances can be more than $200$ words long, which is much longer than average sentence length.}

While most prior research has focused on automatically punctuating monologue speech transcripts, our work is focused on transcripts of real-world phone conversations with turn-taking between two or more people. Spontaneous turn-taking conversations typically are more noisy than monologues because they might contain speech overlaps, false starts, repetitions, and higher transcription word error rates. One way to address this problem is to obtain large amounts of human-punctuated transcripts as training data. However, human annotation is expensive and time-consuming. In this paper, we propose a novel way to leverage freely available external data to improve the punctuation restoration accuracy in human-to-human business conversational data. The main contributions of this paper are as follows:
\begin{itemize}

\item We introduce a new data sampling technique based on an n-gram language model to select additional training data relevant to our target domain from publicly available datasets. Then we propose a two-stage fine-tuning approach based on BERT to train our model.

\item Extensive experiments show that the proposed method is effective to improve performance on unpunctuated noisy data collected from an ASR system. Since our goal is to develop a punctuation restoration model for real-time scenarios in production, we also analyze the performance of different layers in BERT and observe that layer reduction significantly improves the inference speed while maintaining the accuracy. As a secondary contribution, we demonstrate the production environment that we use to deploy the trained model. 

\end{itemize}

\section{Related Work}

In this section, we first review the recent work on punctuation restoration. Since we sample additional training data from external corpora, we also review related work on data sampling.

\subsection{Punctuation Restoration}

 Most of the punctuation restoration models formulate the problem as a sequence labeling task \cite{hentschel2021making, yi2020adversarial, chen2021discriminative, huang2021token, nagy2021automatic}. The model architectures vary from the RNNs to the pre-trained contextualized language models. In recent years, \citet{DBLP:conf/interspeech/TilkA15}, \citet{DBLP:conf/hlt/Salimbajevs16}, and \citet{DBLP:conf/iscslp/XuXY16} used long short-term memory units to restore punctuation in speech transcripts, while \citet{DBLP:conf/iwslt/CourtlandFM20} and \citet{cai2019question} used a bidirectional transformer \cite{DBLP:conf/nips/VaswaniSPUJGKP17} to predict punctuation marks.  

% Another line of study focused on combining prosodic features with lexical features. \cite{DBLP:conf/icassp/Klejch0R17} allows the system to map from per frame acoustic features to word level representations by replacing the traditional encoder in the encoder-decoder architecture with a hierarchical encoder. \cite{DBLP:conf/interspeech/SunkaraRBBK20} explored a multimodal semi-supervised learning approach for punctuation prediction by learning representations from large amounts of unlabelled audio and text data. However, the input to our model is only a transcript, with no access available to acoustic information. 

\subsection{External Training Data Sampling}

%\cite{DBLP:conf/bionlp/ChiuCKP16}, \cite{DBLP:conf/bionlp/KarimiDHN17} select data that is close to the biology doma%. \cite{DBLP:conf/naacl/DaiKHP19} investigates the possibilities of using percentage of vocabulary overlap, language model perplexity, and word vector variance to measure similarity between two datasets.   

For training data sampling, \citet{DBLP:conf/bionlp/ChiuCKP16} and \citet{DBLP:conf/bionlp/KarimiDHN17} used the data that were similar to the biology domain. To measure similarity between two datasets, \citet{DBLP:conf/naacl/DaiKHP19} studied different techniques, such as: vocabulary overlaps, language model perplexity, and word vector variance. They also investigated how similarity measures can be used to sample in-domain pre-training data. Moreover, for NER, they observed that the second phase of in-domain pre-training led to performance gains under both high and low-resource settings.  Recently, \citet{DBLP:conf/emnlp/HanE19} proposed domain-adaptive fine-tuning in which contextualized embeddings were adapted by masked language modeling on texts from the target domain. 

%\citet{DBLP:conf/emnlp/HanE19} proposed domain-adaptive fine-tuning, in which the contextualized embeddings were adapted by masked language modeling on texts from the target domain. \citet{DBLP:conf/naacl/DaiKHP19} showed that the second phase of in-domain pretraining led to performance gains, under both high and low-resource settings. They also investigated how similarity measures can be used to sample the in-domain pre-training data.

\section{Method}

We frame punctuation restoration as a sequence labeling problem. There are four classes that the model needs to predict, including \textbf{PERIOD}\footnote{Although there are few exclamation marks in the data, we convert them to periods for simplicity.}, \textbf{COMMA}, \textbf{QUESTIONMARK}, and \textbf{NONE}.  More specifically, the model predicts what the punctuation mark next to a given token is. If there is no punctuation mark after the input token, the prediction should be \textbf{NONE}. 

\begin{figure*}
\centering
\includegraphics[width=\linewidth]{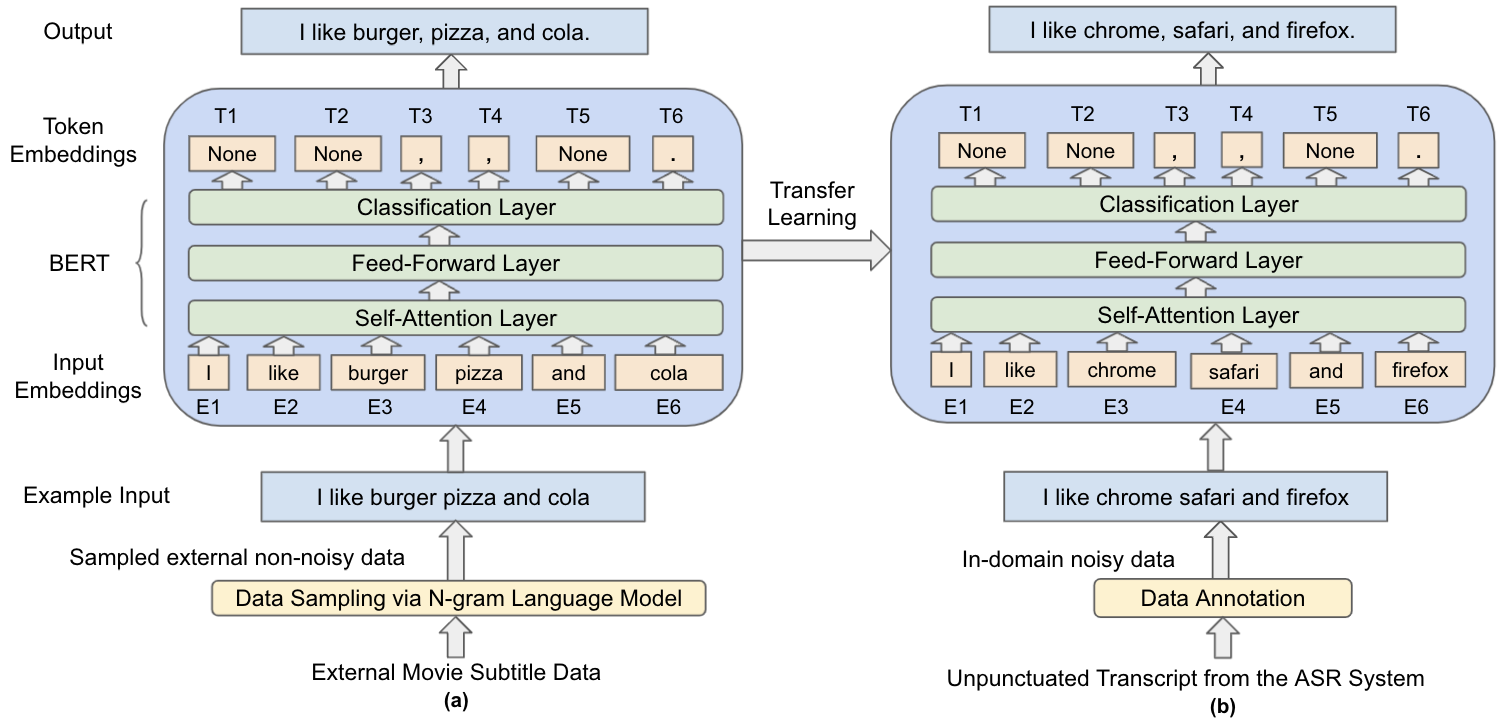}

\caption{Our two stage fine-tuning approach: (a) at first on the external dataset (e.g., movie subtitles), we sample the external data using n-gram similarity and fine-tune the pre-trained BERT model. (b) Next we annotate unpunctuated transcripts generated from an ASR system to construct the in-domain training set (noisy data) and apply the second stage of fine-tuning using the BERT model fine-tuned in the previous stage.}
\label{fig:model_overview}
\end{figure*}

In Figure~\ref{fig:model_overview}, we show the overall architecture of our proposed model that utilizes a two stage fine-tuning approach: First, (i) we do data sampling from an external dataset (e.g., movie subtitles) using our proposed n-gram similarity approach and fine-tune a pre-trained BERT model. Then, (ii) we annotate unpunctuated phone conversation transcripts generated from an ASR system to construct an in-domain training set (noisy data) to apply the second stage of fine-tuning using the BERT model fine-tuned in the previous stage. Below, we describe the data sampling, data annotation, and two-stage fine-tuning approach in detail.

\nop{
\begin{table*}[t]
\begin{tabular}{|c|c|}
\hline
\textbf{Out-of-domain (External Dataset)} & \textbf{In-domain (Noisy Phone Conversations)} \\ \hline
Our dog disappeared. & Well, thank you for coming down. \\
Sometimes those serious ones fool you. & What did you find out so far? \\
He looks so utterly vulnerable. & I guess you could give me her number. \\
God destroyed whole cities to punish man's wickedness. & Um, just hold on a second. \\
I mean, who comes hereafter dark besides murderers? & Might I have your email? \\ \hline
\end{tabular}
\caption{Some examples from the external and the in-domain datasets.}
\label{tab:domains}
\end{table*}
}

\subsection{Data Sampling from the External Dataset via N-gram Language Model}

In principle, we can obtain a tremendous amount of punctuated text from the internet (e.g English Wikipedia). However, written English is fundamentally different from spoken English and training a model on a corpus that is distant from the target domain might end up harming model performance. Thus, we choose to use a movie subtitle corpus, OpenSubtitles \cite{DBLP:conf/lrec/LisonT16}, as our first stage fine-tuning data because there are many human-to-human conversations in movies. 

As movie conversations are not necessarily business conversations, training a model on out-of-domain data might also degrade model accuracy. Thus, we propose a data sampling method based on language model perplexity to select data similar to our target domain (noisy phone conversations). Some examples from the out-of-domain (movie subtitles) data and the in-domain (noisy phone conversations) data are shown in Table \ref{tab:domains}. 

For data sampling, we follow \cite{DBLP:conf/naacl/DaiKHP19} and assign a probability to any sequence of words $<w_1, \cdots, w_N >$ using the n-gram language model \cite{jurafskyspeechngram}. The probability of a sequence can be obtained by chain rule of probability. 
\begin{equation}
\label{eq:lm}
p(w_1, w_2, \cdots, w_N)=\prod_{i=1}^N p(w_i|w_1^{i-1})
\end{equation}

Here, $N$ is the length of the sequence and $w^{i-1}_1$ are all words before word $w_i$. In our work, we use a 4-gram language model. 

To sample data that are similar to our in-domain data, we first train a language model on our target domain, which is business phone conversations. Then we evaluate the utterances (subtitles) in the movie subtitle corpus based on the perplexity generated by the model for a given utterance (subtitle) to represent the degree of similarity. The main difference between \cite{DBLP:conf/naacl/DaiKHP19} and our work is that we measure the similarity at the utterance level instead of the corpus level. When an utterance is unlikely to occur in the data where the language model is trained on, it will assign a low probability and high perplexity. More specifically, we feed an utterance from the movie subtitle corpus to the 4-gram language model and then the model estimates its perplexity. Afterwards, we rank the subtitles by perplexity and select the top 4.8M subtitles as analogous to in-domain data. Some examples from the external and the in-domain data are shown in Table ~\ref{tab:domains}.

\begin{table}[t]
\small
\centering
\begin{tabular}{|c|}
\hline
\textbf{Out-of-domain (External Movie Subtitle Dataset)} \\ \hline
Our dog disappeared.  \\
Sometimes those serious ones fool you.\\
He looks so utterly vulnerable. \\
God destroyed whole cities to punish man's wickedness. \\
I mean, who comes hereafter dark besides murderers? \\
Your paintings were impressive joker. \\
\hline
\textbf{In-domain (Noisy Phone Conversations)} \\ \hline
Well, thank you for coming down. \\
 What did you find out so far? \\
 I guess you could give me her number. \\
  Um, just hold on a second. \\
  Might I have your email? \\ 
  Eh, I really don't know what to do about it! \\ 
\hline
\end{tabular}
\caption{Examples from the out-of-domain and the in-domain datasets.}
\label{tab:domains}
\end{table}

\begin{table}[t]
\centering
\begin{tabular}{|c|c|c|c|}
\hline
\textbf{Data type} & \textbf{Comma} & \textbf{Period} & \textbf{Question} \\ \hline
Training & $1166386$  & $860403$  & $118203$  \\ \hline
Dev      & $15899$    & $14143$   & $1722$    \\ \hline
Test     & $6503$     & $3880$    & $629$     \\ \hline
\end{tabular}
\caption{In-domain dataset (noisy data) distribution.}
\label{tab:datadistribution}
\end{table}

\begin{table*}[t]
\centering
\begin{tabular}{|c|c|c|c|c|c|}
\hline
\textbf{Model} & \textbf{First fine-tuning} & \textbf{Second fine-tuning} & \textbf{Precision} & \textbf{Recall} & \textbf{F1 score} \\ \hline
Indom & In-domain data & $\times$ & 75.01 & 63.71 & 68.48 \\ \hline
Samex & Sampled external data & $\times$ & 67.70 & 57.49 & 61.68 \\ \hline
Two-stage & Sampled external data & In-domain data & 75.94 & 65.18 & 69.60 \\ \hline
\end{tabular}
\caption{Performance based on different fine-tuning techniques with 6-layer BERT.}
\label{tab:benchmarking}
\end{table*}
\subsection{Data Annotation for the ASR Data}

We sample 320,000 utterances produced by the ASR model from business phone calls across a time period of a year (see Table \ref{tab:datadistribution} for details). We also make sure utterances are distributed homogeneously among different scenarios. More importantly, we follow our standard data privacy protocol while working on data annotation. Since human-to-human conversations are incredibly noisy that could degrade the performance of a pre-trained language model that is trained on written texts, we remove repetitions, false starts, and filler words from transcripts to address this problem. Note that inserting punctuation marks to unpunctuated texts might take a long time for an annotator. To speed up this process, we use a punctuation restoration model based on BERT \cite{DBLP:conf/naacl/DevlinCLT19} that is trained on a much smaller dataset to produce default punctuation marks for utterances and let the annotators edit these punctuation marks. In total, 102 annotators contributed to the annotation job.

\subsection{Two-Stage Fine-tuning Approach}

Fine-tuning a pre-trained language model at first in a dataset similar to the target domain, and then fine-tuning it again in the target dataset is found effective in different tasks, such as answer selection \cite{DBLP:conf/aaai/GargVM20}, and query focused text summarization \cite{DBLP:conf/ai/LaskarHH20,DBLP:conf/coling/LaskarHH20}. In this paper, we also utilize a two-stage fine-tuning approach that transfers a pre-trained language model to the punctuation restoration task in the first-stage; and then adapts the obtained model to the specific target domain in the second-stage. %Note that we select the business domain as our target domain. 
For that purpose, we use the BERT model \cite{DBLP:conf/naacl/DevlinCLT19} that is a Transformer-based \cite{DBLP:conf/nips/VaswaniSPUJGKP17} contextual language model as our backbone model. To utilize BERT for sequence labeling, a classification layer is added on top of the last layer (see Figure \ref{fig:model_overview}). In addition, as our goal is to deploy a punctuation restoration model in real-time and at scale, we use the weights of the first (bottom) six layers of the BERT base \cite{DBLP:conf/naacl/DevlinCLT19} model (12 transformer layers), to initialize a 6-layer BERT model for fine-tuning.
%, and then fine-tune the 6-layer model. %Note that we also compare the performance of BERT when other different layers are used.

\nop{
\subsection{Layer Reduction}

As our goal is to run a punctuation restoration model in real-time and at scale, we explore the possibility of training a smaller BERT model with little to no loss in accuracy. The inference time of a Transformer-based model is approximately proportional to the	number of transformer layers. Reducing the number of layers could significantly improve the inference time. We compare models with different number of layers.

To reduce the number of layers, we use the weights of the first bottom six layers of the BERT base model, which has 12 transformer layers in total, to initialize a 6-layer BERT model, and then fine-tune the 6-layer model.
}

\begin{table}[tbh]
\centering
\begin{tabular}{|c|c|c|c|}

\hline
\textbf{Type} & \multicolumn{1}{c|}{\textbf{Precision}} & \multicolumn{1}{c|}{\textbf{Recall}} & \multicolumn{1}{c|}{\textbf{F1}} \\ \hline
Period        & 72.57      & 77.89       & 75.14        \\ \hline
Question      & 84.63      & 63.91       & 72.83        \\ \hline
Comma         & 77.23      & 57.79       & 66.11        \\ \hline
Overall       & 75.94      & 65.18       & 69.60         \\ \hline
\end{tabular}
\caption{Performance by the Two-stage model on different punctuation marks.}
\label{tab:twostage}
\end{table}

\section{Experiments}

In our experiments, we investigate the following: \textit{(i) Will the model perform better if we use external datasets?} \textit{(ii) Will the model performance degrade if we reduce the number of layers?} Below, we first describe our experimental settings. Then we analyze the model performance when external data are used, followed by discussing the performance when layer reduction is applied. 

\subsection{Experimental Settings}

To fine-tune 6 layer BERT models, we used the following hyperparameters: gradient accumulation steps = $1$, warmup proportion = $0.1$, learning rate = $5e-5$, train batch size = $50$, and the max sequence length = $300$. All training experiments were run in an Google Cloud Platform (GCP) instance using 1 NVIDIA P100 GPU. To investigate the inference speed for real-time scenarios, we ran our experiments with $1$ CPU, Intel(R) Xeon(R) CPU@$2.20$GHz. 

\subsection{Performance while using External Data}

We fine-tune models on various data settings and evaluate them on the test set. In the first data setting, we fine-tune the BERT model only on the in-domain data. We denote this model as Indom. In the second setting, the BERT model, fine-tuned only on the sampled external data is denoted as Samex. The last setting is our proposed Two-stage approach that first fine-tunes BERT on the sampled external data and then fine-tunes again on the in-domain data. The results are shown in Table~\ref{tab:benchmarking}.

The model (Samex) that was trained only on sampled external data unsurprisingly shows the F1 score as 61.68, which is 7.92\% lower than the model (Two-stage) that was also fine-tuned on the in-domain data. This indicates that in-domain fine-tuning is crucial to boost model performance in a target domain. We also find that the model (two-stage) fine-tuned on both the in-domain and the sampled external data outperforms the model (Indom) fine-tuned only on the in-domain data by 1.12\% F1 score. This demonstrates the effectiveness of two-stage fine-tuning using external data.

Table~\ref{tab:twostage} shows how our best performing model (two-stage) performs in three different punctuation marks with the 6-layer BERT model. It is worth noting that the recall of commas is much lower than other punctuation types. One explanation is that although we tried our best to remove repetitions, false starts, and filler words, there are still a noticeable amount of them in the transcripts. Thus, it often makes it very challenging to place commas. 
    
\nop{
\begin{table}[tbh]
\begin{tabular}{|c|c|c|c|}
\hline
\textbf{Model} & \textbf{Precision} & \textbf{Recall} & \textbf{F1 score} \\ \hline
3-layer BERT & 73.49 & 62.93 & 67.25 \\ \hline
6-layer BERT & 75.01 & 63.71 & 68.48 \\ \hline
12-layer BERT & 74.27 & 64.09 & 68.36 \\ \hline
\end{tabular}
\caption{Performance based on the number of layers.}
\label{tab:layerreduction}
\end{table}
}

\begin{table}[tbh]
\begin{tabular}{|c|c|c|c|c|}
\hline
\textbf{Model} & \textbf{P} & \textbf{R} & \textbf{F} & \textbf{Inf. time}  \\ \hline
3-layer & 73.49 & 62.93 & 67.25 & 13.83 \\ \hline
6-layer & 75.01 & 63.71 & 68.48 & 26.52 \\ \hline
12-layer & 74.27 & 64.09 & 68.36 & 50.01 \\ \hline
\end{tabular}
\caption{Performance based on different number of BERT layers. Here, Precision, Recall, and F1 are denoted by P, R, and F respectively; while the Inference Time (Inf. time) is based on milliseconds per example.}
\label{tab:layerreduction}
\end{table}

\subsection{Performance based on Layer Reduction}
To know the effect of layer reduction in BERT, we fine-tune three BERT models with different numbers of layers (3, 6, 12) on the in-domain data only. This setting is similar to the Indom model except changing the number of layers. The results are shown in Table~\ref{tab:layerreduction}. It is worth noting that there is no significant difference in model performance between the smaller models (3-layer and 6-layer) and the 12-layer model. The 6-layer model even outperforms the 12-layer model in terms of Precision and F1 scores. More importantly, the inference time is significantly decreased by 47.96\% and 72.35\% from the 12-layer model to the 6-layer and the 3-layer models respectively.

\subsection{Production Deployment}
For production inference, we also remove the filler words and repetitions similar to the training phase  to ensure that the input distributions are similar. For production deployment, we use an Intel(R) Xeon(R) CPU@2.20GHz and feed one utterance to the model at once (i.e., batch size = 1). The model runs in real-time in our production environment. The average inference speed is $201$ milliseconds per utterance.

\section{Conclusion}

In this paper, we propose a new approach to sample training data that are similar to the target domain from freely available general corpora via an n-gram language model. We conduct extensive experiments using BERT with different data settings for the punctuation restoration task and find that our two-stage fine-tuning approach is effective to improve model performance. We also show the effectiveness of our model for real-time inference as we find that removing the first 6 layers of the 12 layer BERT does not harm the accuracy but substantially improves the speed. In the future, we will investigate how other external datasets \cite{dblp:chen2021summscreen} impact the model performance.

\section*{Acknowledgements} We thank Harsh Saini for help with model deployment, Shayna Gardiner for giving ideas about the writing, and Feriba Saboor for help with data annotation. %We would also like to thank other members in the Natural Language Processing (NLP) team, the Automatic Speech Recognition (ASR) team, and the Data Engineering (DE) team at Dialpad for their help in this work.

% \balance
\bibliography{anthology,custom}
\bibliographystyle{acl_natbib}

\appendix

\end{document}